\pdfoutput=1

\documentclass[11pt]{article}

\usepackage[preprint]{acl}

\usepackage{times}
\usepackage{latexsym}

\usepackage[T1]{fontenc}

\usepackage[utf8]{inputenc}

\usepackage{microtype}

\usepackage{inconsolata}

\usepackage{graphicx}

\usepackage{cleveref}
\usepackage{tabularx}
\usepackage{booktabs}
\usepackage{xcolor}
\usepackage{fancyvrb}

\title{Fact Finder - Enhancing Domain Expertise of Large Language Models by Incorporating Knowledge Graphs}

\author{
    Daniel Steinigen\textsuperscript{1} \\ 
    {\bf Roman Teucher\textsuperscript{1}} \\ 
    {\bf Timm Heine Ruland\textsuperscript{1}} \\ 
    {\bf Max Rudat\textsuperscript{1}} \\ 
    {\bf Nicolas Flores-Herr} \\
    Fraunhofer IAIS \\
    Sankt Augustin, Germany \\
    \small{\texttt{<first-name>.<last-name>@iais.fraunhofer.de}} \\
    \small{\textsuperscript{1}All authors contributed equally to this work} \\\And
    Peter Fischer\textsuperscript{2}\\ 
    {\bf Nikola Milosevic\textsuperscript{2}} \\ 
    {\bf Christopher Schymura\textsuperscript{2}}  \\ 
    {\bf Angelo Ziletti\textsuperscript{2}} \\
    Bayer AG \\
    \small{\texttt{<first-name>.<last-name>@bayer.com}} \\
    \small{\textsuperscript{2}Shared senior authorship} \\
}

\begin{document}
\maketitle
\begin{abstract}
Recent advancements in Large Language Models (LLMs) have showcased their proficiency in answering natural language queries. However, their effectiveness is hindered by limited domain-specific knowledge, raising concerns about the reliability of their responses. We introduce a hybrid system that augments LLMs with domain-specific knowledge graphs (KGs), thereby aiming to enhance factual correctness using a KG-based retrieval approach. We focus on a medical KG to demonstrate our methodology, which includes (1) pre-processing, (2) Cypher query generation, (3) Cypher query processing, (4) KG retrieval, and (5) LLM-enhanced response generation. We evaluate our system on a curated dataset of 69 samples, achieving a precision of 78\% in retrieving correct KG nodes. Our findings indicate that the hybrid system surpasses a standalone LLM in accuracy and completeness, as verified by an LLM-as-a-Judge evaluation method. This positions the system as a promising tool for applications that demand factual correctness and completeness, such as target identification --- a critical process in pinpointing biological entities for disease treatment or crop enhancement. Moreover, its intuitive search interface and ability to provide accurate responses within seconds make it well-suited for time-sensitive, precision-focused research contexts. We publish the source code together with the dataset and the prompt templates used\footnote{\url{https://github.com/chrschy/fact-finder}}.
\end{abstract}

\begin{figure*}[!htb]
    \begin{center}
        \includegraphics[width=1.0\textwidth]{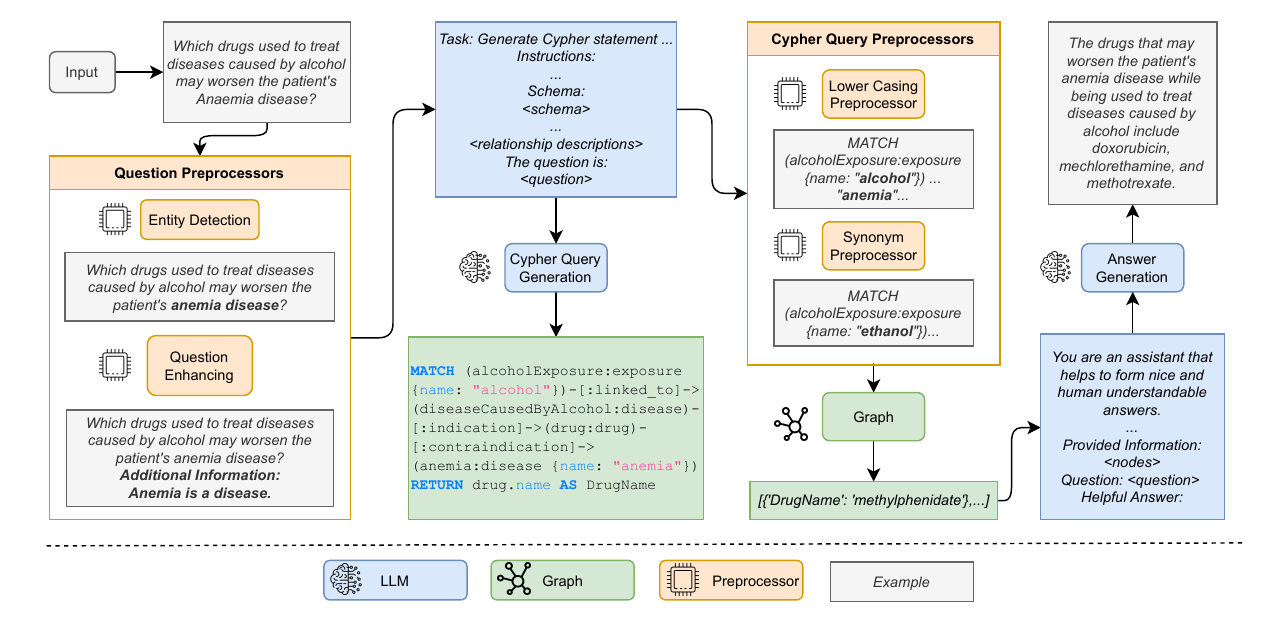}
        \caption{Overview of the FactFinder pipeline using large language models and knowledge graphs to answer scientific questions.}
        \label{fig:flowchart}
    \end{center}
\end{figure*}

\section{Introduction}
Recently, Large Language Models (LLMs) have enabled sophisticated question-answering systems, revolutionizing the landscape of natural language processing \cite{openai2023gpt4,jiang2023mistral,bubeck2023sparks,park2023generative,touvron2023llama,katz2023gpt}. These advanced models, with their ability to understand and generate human-like text, have shown great potential in various domains, including life sciences \cite{nori2023capabilities,waisberg2023gpt,bavsaragin2024you}. 
However, LLMs are limited by the timeframe of their training data and can produce incorrect statements, known as hallucinations~\cite{ziwei2023}, or incomplete answers by providing only a few relevant entities while missing others not included in their internal knowledge.

In domains such as life sciences, obtaining answers with current and factual information is paramount for many use cases \cite{malaviya2023expertqa,ljajic2024scientific}. Factually correct AI-generated reviews can aid researchers in information retrieval and hypothesis building. 
For example, target identification requires up-to-date knowledge of the latest literature. Target identification involves pinpointing a biological entity, such as a gene or protein, that can be manipulated to achieve a desired effect, like treating a disease in humans (pharmaceuticals) or improving crop resilience in plants (crop sciences). 
Similarly, designing effective field or clinical trials requires considering up-to-date information. For field trials in crop sciences, this may include environmental and climatic conditions, market developments, and regulatory requirements. For clinical trials and medical writing, relevant information includes details about the drug under development, market conditions, planned sites, and regulatory requirements. 
Current and timely information is crucial for competitive intelligence, including insights on competitor products, disease epidemiology, and market size. AI-based solutions can assist in identifying concurrent disease occurrences and help researchers develop hypotheses using real-world data.

Knowledge Graphs (KGs) represent a promising strategy for improving factual correctness in LLMs \cite{jiang-etal-2023-structgpt,baek-etal-2023-knowledge, sen-etal-2023-knowledge}, including the life science domain~\cite{feng2024-Knowledge}. By organizing entities such as drugs, diseases, and genes, along with their relationships, into a structured network, KGs provide useful additional context for LLMs for precise and relevant information retrieval \cite{chandak2023building,milovsevic2023comparison,badenes2023-question}. This organized data framework allows LLMs to produce more factually accurate and comprehensive responses~\cite{pan2024unifying}.
In addition, KGs enable systems to leverage current and comprehensive information, including recent data not available during the LLMs' training phase. From an organizational standpoint, integrating KGs enhances top-tier LLMs with proprietary or specialized knowledge. This integration facilitates the inclusion of unique organizational data sources, such as historical and ongoing lab experiments or licensed datasets. 
 
In this paper, we present FactFinder - a hybrid question answering (QA) system - which leverages both KG and LLM to provide answers to scientific questions. Fig.~\ref{fig:flowchart} depicts the system's architecture, which is structured as a pipeline with several subcomponents. 
Our main contributions are:
\begin{itemize}
\item We provide an easy-to-use system that answers scientific questions combining LLMs and KGs.
\item We release a dataset of manually annotated text-to-Cypher query pairs, which could serve as benchmark for validating text-to-Cypher conversion system.
\item We present a methodology showing that current state-of-the-art LLMs are able to generate satisfactory Cypher queries for the life science domain.
\item We share our dataset, source code, and prompt templates\footnote{\url{https://github.com/chrschy/fact-finder}}.
 \end{itemize}

\section{Data}
\noindent \textbf{Knowledge graph.}
We use PrimeKG \cite{chandak2023building} as our source of fact-based background knowledge. PrimeKG integrates 20 high-quality resources to describe 17,080 diseases with 4,050,249 relationships, including over 100,000 nodes and 29 types of edges that densely connect disease nodes with drugs, genes, exposures, and phenotypes.
We preprocess the graph data by mapping names to their preferred terms, as described in \Cref{ssec:querygen}, and converting all entries to lowercase.

\noindent \textbf{Text-to-Cypher dataset.}
\label{ssec:text-to-cypher-data}
We manually generated a ground-truth dataset containing 69 text-to-Cypher query pairs specifically designed for medical questions. These queries are complex, often involving multiple hops in the graph, aggregation, and boolean question structures. They require deep knowledge of Cypher and the KG. Each entry includes a natural language question, the corresponding Cypher query, the expected answer, and relevant nodes and relationships.
This dataset provides a benchmark to evaluate the ability of text-to-Cypher systems to interpret and execute complex queries. While specialized to the PrimeKG graph, the dataset leverages PrimeKG's extensive applicability, making it a valuable resource for various medical information retrieval tasks.
Examples from the dataset include simpler questions like \textit{Which drugs have pterygium as a side effect?} and more complex ones such as \textit{Which medications have more off-label uses than approved indications?} and \textit{Which diseases have only treatments that have no side effects at all?}

\section{System Description}

\subsection{Cypher Query Generation}
\label{ssec:querygen}
Generating code to query structured databases from natural language inquiries used to be a complex process, involving steps such as entity and relation extraction, entity and relation linking, query type classification, template-based or compositional query generation \cite{srivastava-etal-2021-complex,https://doi.org/10.1002/widm.1389}.
Retrieval-based approaches like Graph-RAG aim to simplify this by partitioning the graph into communities of nodes and edges, which are then retrieved and summarized using a LLM to generate an answer \cite{edge2024local}. However, Graph-RAG struggles with complex queries that span multiple graph communities

With the advent of LLMs, however, QA systems can now understand domain-specific questions and generate valid queries directly, allowing for more flexible approaches.
Much of the research has been centered on text-to-SQL generation, where LLMs have demonstrated considerable effectiveness \cite{gao-2023-text,chang2023prompt}, including in the medical domain \cite{ziletti2024retrieval}, and have often performed better than specialized models \cite{pourreza-2023-dinsql}.
Conversely, the area of text-to-Cypher query generation remains relatively under-explored, with prior research primarily focused on sequence-to-sequence models \cite{zhao2024-cyspider,guo2022-spsql}. Only recently has the application of LLMs to this task begun to emerge \cite{feng2023-robust}.
To bridge this gap, our work evaluates the capabilities of LLMs to produce Cypher queries for scientific QA in the medical domain (see Sec.~\ref{ssec:evalgraph}). 

We prompt LLMs with questions and graph schemas, including node and relationship types and their properties, to generate Cypher queries.
When a graph relation is not self-explanatory, we add its natural language description to the prompt. For instance, for the relation \textit{ppi}, we add \textit{"Temporary, non-covalent binding between protein molecules. Protein-protein interactions occur..."}.
During instruction prompting, we also identify questions that cannot be answered by the given graph schema. In such cases, the LLM returns the string SCHEMA\_ERROR along with a brief explanation of why it could not generate an answer. FactFinder detects this marker using a regex and returns the explanation to the user.

We include an entity extraction model to align entity names in questions with those in the KG, based on Linnaeus~\cite{gerner2010-linnaeusas} and developed set of vocabularies for entity types. This step ensures consistency by replacing detected entities with their preferred KG terms (e.g., \textit{alcohol} to \textit{ethanol}) and generating sentences linking each entity to its category (see Fig.~\ref{fig:flowchart} left), reducing the LLM's reliance on domain-specific knowledge.

\subsection{Query Pre-Processors}
Before querying the graph with the generated Cypher query, we preprocess it to increase the system's robustness. This leverages the structured natural language understanding provided by the translation of questions to Cypher queries. Various regular expression-based methods target specific query elements.

\noindent \textbf{Formatting.} First, we format the query to improve readability and consistency. This includes adding indentation, line breaks, and ensuring consistent naming conventions, which simplifies the application of regular expressions in subsequent steps.

\noindent \textbf{Lowercasing Property Values.} We convert property values in the Cypher query to lowercase, matching the pre-lowered graph properties. 

\noindent \textbf{Synonym Selection.} We map entities in the Cypher query to preferred terms used in the graph. If no mapping is available, we use external tools (e.g., \textit{skos:altLabel} queries against Wikidata) to find synonyms and match them to graph terms.

\noindent \textbf{Deprecated Code Handling.} We correct deprecated code generated by the LLM, such as replacing the obsolete SIZE() keyword with the current COUNT() keyword.

\noindent \textbf{Child to Parent Node Mapping.} In PrimeKG~\cite{chandak2023building}, some node types are connected by parent-child relationships. We replace child nodes with their parent nodes in the Cypher query to ensure completeness.

\subsection{Graph Question Answering and Verbalization}

The pre-processed Cypher query is executed on the graph, returning a unique set of nodes that may include names, properties, IDs, and other elements.
Next, the question and graph results are incorporated into a prompt template and sent to an LLM. The prompt instructs the LLM to rely solely on the graph information to formulate an answer, which is then provided as the final natural language output.
We use the Neo4j graph database\footnote{\url{https://neo4j.com/}} to provide the KG and build our pipeline borrowing components from Langchain\footnote{\url{https://www.langchain.com/}}.

\subsection{Explainability through Evidence}
\label{ssec:evidence}


\begin{figure}[htb]
\centering
\includegraphics[width=0.45\textwidth]{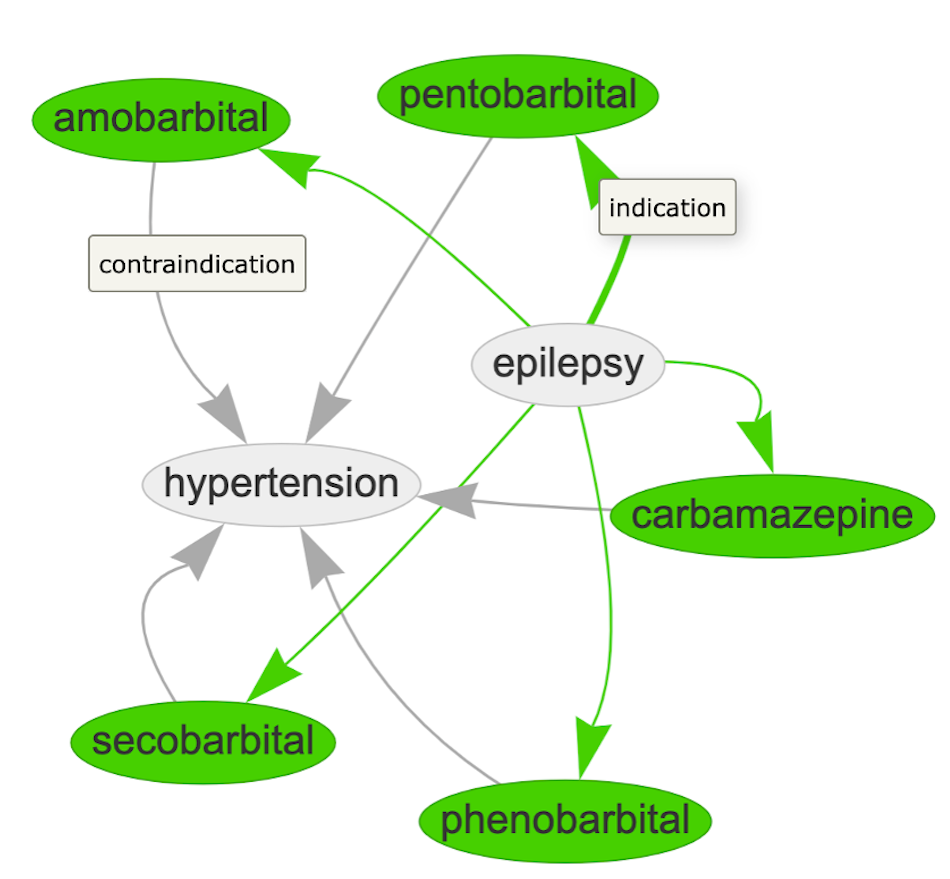}
\caption{Example of the evidence subgraph for \textit{Which drugs against epilepsy should not be used by patients with hypertension?}}
\label{fig:screenshot-graph}
\end{figure}

To ensure transparency and explainability, the system provides various forms of evidence alongside the natural language answer. These include intermediate results, such as Cypher generation prompts and graph results, as well as explicitly created information like the underlying subgraph for a query. These evidences are displayed in the frontend to track the system's behavior.

\noindent \textbf{Cypher Query Evidence.} The primary evidence is the Cypher query generated by the model, demonstrating how the given question maps to the graph structure. This enables expert users to evaluate the model's understanding of the question and the quality of the generated query.

\noindent \textbf{Graph Response.} The system also provides the actual response from the graph, consisting of the nodes and relationships returned by the executed query.

\noindent \textbf{Subgraph Visualization.} To enhance interpretability, we provide a subgraph as part of the evidence. This subgraph visually displays (via Pyvis\footnote{\url{https://pyvis.readthedocs.io/}}) the relevant nodes and edges, illustrating the subset of the main graph that contributed to the specific answer, as shown in Fig.~\ref{fig:screenshot-graph}. 

\noindent \textbf{Sub-Graph Generator.} A notable challenge is that the original Cypher query usually returns only the nodes required to answer the question, omitting the edges that connect the question's entity to the answer nodes. To address this, we generate a new Cypher query that fetches both the answer nodes and the connecting edges. This is achieved by submitting the original Cypher query to a LLM and instructing it to return all nodes and edges present in the query.

\noindent For example, if our original Cypher query is:
\begin{Verbatim}[commandchars=\\\{\}]
 \textcolor{cyan}{MATCH} (g:gene\_or\_protein {\textcolor{cyan}{name}:\textcolor{magenta}{"pink1"}})-
 [:associated\_with]->(d:disease) 
 \textcolor{cyan}{RETURN} d.id \textcolor{cyan}{AS} ID, d.\textcolor{cyan}{name} \textcolor{cyan}{AS} Name
\end{Verbatim}
our subgraph Cypher query is:
\begin{Verbatim}[commandchars=\\\{\}]
 \textcolor{cyan}{MATCH} (g:gene\_or\_protein {\textcolor{cyan}{name}:\textcolor{magenta}{"pink1"}})-
 [a:associated\_with]->(d:disease) 
 \textcolor{cyan}{RETURN} g, d, a
\end{Verbatim}

\noindent Note that this last subgraph Cypher query returns all relevant nodes and edges.

\subsection{User Interface and Example of Usage}

\begin{figure}[htb]
\centering
\includegraphics[width=0.45\textwidth]{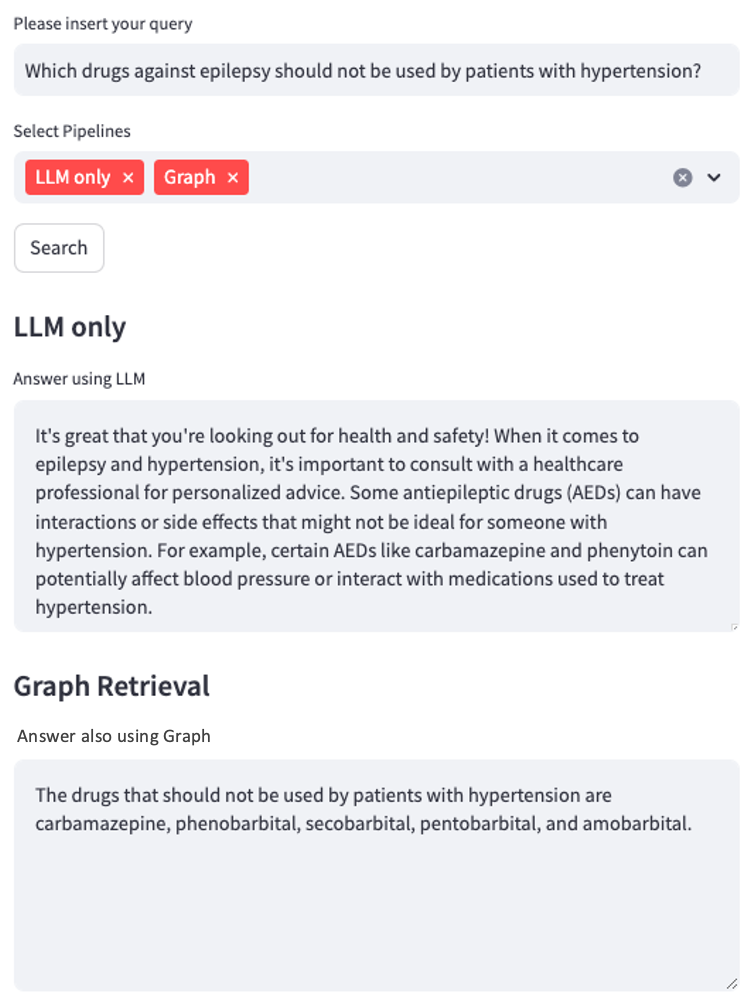}
\caption{User Interface with question and answers of the standalone LLM and our graph-based hybrid system.}
\label{fig:screenshot-answers}
\end{figure}

\begin{figure}[htb]
\centering
\includegraphics[width=0.45\textwidth]{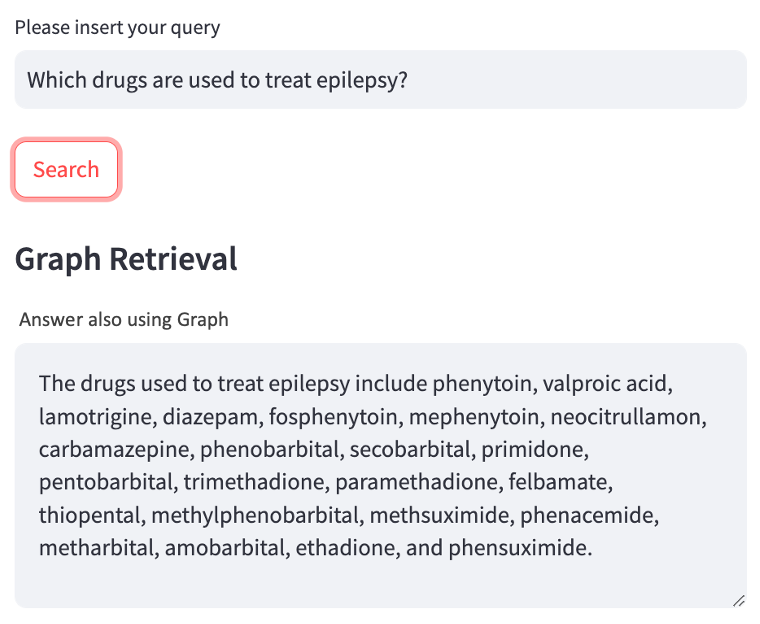}
\caption{Answer for exploring drugs used to treat epilepsy.}
\label{fig:screenshot-drugs}
\end{figure}

\begin{figure}[htb]
\centering
\includegraphics[width=0.45\textwidth]{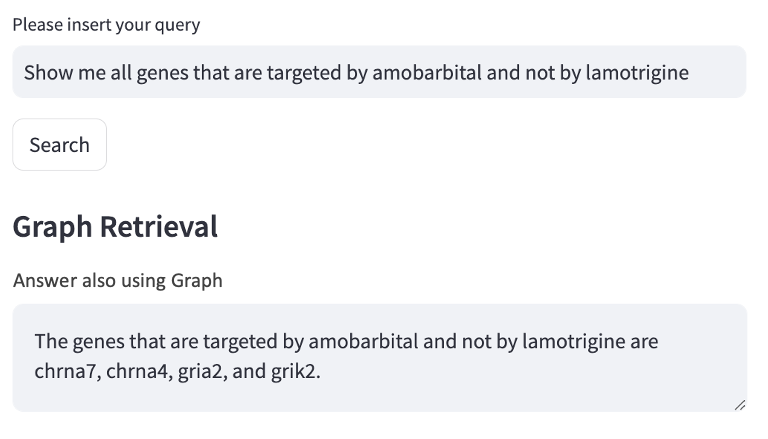}
\caption{Answer for exploring genes targeted by amobarbital but not lamotrigine.}
\label{fig:screenshot-genes}
\end{figure}

\begin{figure*}[htb]
\begin{center}
\includegraphics[width=1.0\textwidth]{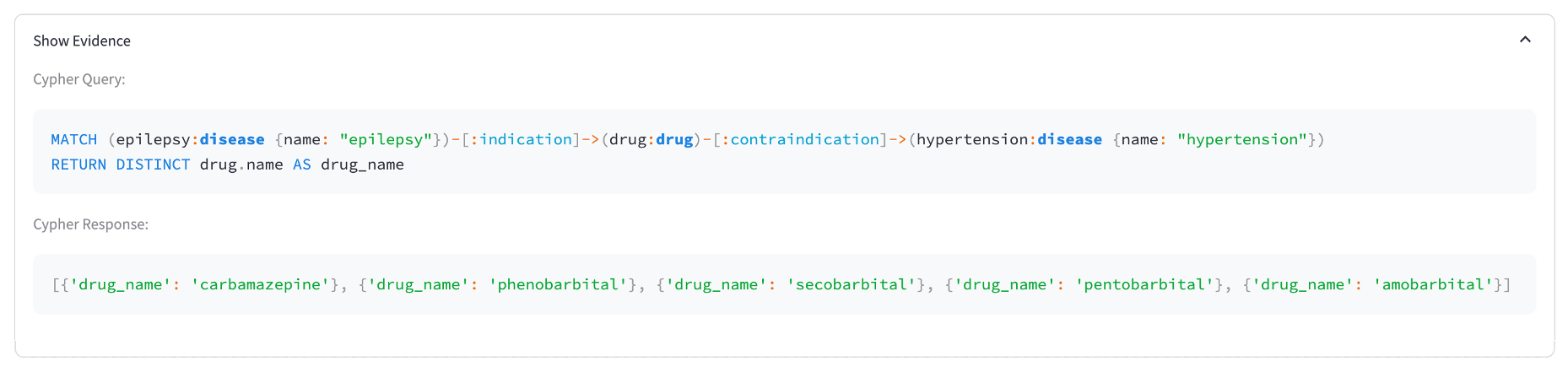}
\caption{User Interface with the generated Cypher query and the graph response as evidence.}
\label{fig:screenshot-evidence}
\end{center}
\end{figure*}
Our target audience includes researchers in the life sciences, such as those working in medical research and crop science, who are interested in exploring connections between drugs, genes, proteins, and other biological entities to discover new research directions.
To make our pipelines accessible, we developed a graphical user interface using Streamlit\footnote{\url{https://streamlit.io/}}. This interface allows users to input questions and view generated answers. Users can select different pipelines, such as LLM-only or those incorporating KGs or documents.
For instance, a medical researcher might ask, \textit{Which drugs against epilepsy should not be used by patients with hypertension?} (Fig.~\ref{fig:screenshot-answers}). To delve deeper, they could follow up with \textit{Which drugs are used to treat epilepsy?} (Fig.~\ref{fig:screenshot-drugs}).
Going further, with the question \textit{Which genes are targeted by amobarbital but not lamotrigine?}, they could take a first step towards understanding the genetic interactions that differentiate drug effects, resulting in the identification of four genes exclusively targeted by amobarbital (Fig.~\ref{fig:screenshot-genes}).
Additionally, users can interactively visualize the relevant subgraph, generated Cypher query, and graph response, enabling them to verify response accuracy and understand the underlying data (Fig.~\ref{fig:screenshot-evidence}). These features (see \Cref{ssec:evidence}) enhance transparency and foster trust in the system.

\section{Evaluation}
\label{sec:eval}
\subsection{Graph Retrieval Evaluation}
\label{ssec:evalgraph}
To quantify the graph retrieval step, we assess the returned nodes from graph queries by comparing result nodes from executing the ground truth queries with those from the generated queries. This enables a quantitative evaluation of the text-to-Cypher step. We use the ground truth text-to-Cypher dataset described in Sec. \ref{ssec:text-to-cypher-data} for this evaluation. We compute intersection over union (IoU), precision, and recall for the expected and generated graph result sets, as shown in Table \ref{tab:eval-retrieval}.

The results indicate strong performance, with the best model exceeding 75\%. GPT-4o outperforms GPT-4-Turbo overall. Entity Enhancement (EE, Fig.~\ref{fig:flowchart} left and Sec.~\ref{ssec:querygen}) improves GPT-4-Turbo's performance but slightly decreases GPT-4o's effectiveness. Manual analysis revealed that PrimeKG's merging of genes and proteins into a single node can mislead GPT-4o when EE is applied, directing it towards incorrect relations. GPT-4o's internal knowledge allows it to infer relationships more accurately without EE, while GPT-4-Turbo benefits from the additional clarity provided by EE. 
This suggests that LLM's internal knowledge can be beneficial for Cypher query generation.
%
\begin{table}[!ht]
\centering
\begin{tabular}{@{}lcrrr@{}}
\toprule
\textbf{Model} & \textbf{EE} & \textbf{IoU} & \textbf{Precision} & \textbf{Recall} \\ 
\midrule
 gpt-4-turbo & True & 71.3 & 73.6 & 71.6 \\  
 gpt-4-turbo & False & 62.7 & 65.3 & 64.9 \\
 gpt-4o & True & 74.9 & 77.0 & 77.6 \\
 gpt-4o & False & \textbf{75.2} & \textbf{77.5} & \textbf{77.8} \\
\bottomrule
\end{tabular}
\caption{Results for the graph retrieval evaluation (metrics in \%). IoU stands for intersection over union and EE for Entity Enhancement, see \Cref{ssec:evalgraph}.}
\label{tab:eval-retrieval}
\end{table}

\subsection{Evaluating Correctness and Completeness}
\label{ssec:evaljudge}
To assess the quality of LLM-generated answers, we conduct two evaluations using the LLM-as-a-Judge approach~\cite{zheng2024judging}: (1) comparing the answers from our KG-LLM-based system to those from an LLM-only system, and (2) evaluating the reliability of LLM verbalization of information provided by the KG. In both cases, correctness is defined as the inclusion of only facts from the graph nodes, and completeness as the inclusion of all such facts. 

\noindent \textbf{Hybrid system vs. LLM-only.} 
We compare the hybrid KG-based system against a standalone LLM. The hybrid system (GPT-4o without entity enhancement, Sec.~\ref{ssec:evalgraph}) is evaluated to produce more correct (complete) answers in 94.12\% (96.08\%) of cases, demonstrating its superior performance in providing accurate and complete responses.

\noindent \textbf{LLM verbalization.} 
We evaluate the verbalization of natural language answers from graph results. In this evaluation, 89.13\% of answers are deemed correct, and 80.43\% complete, indicating high accuracy in verbalization.

%
%
\subsection{Handling Incorrect Graph Responses}
Finally, we evaluate FactFinder's ability to handle incorrect or incomplete information in graph responses. The system should be able to refuse to answer if the Cypher query generation step produces a wrong query, thus retrieving the correct data from the KG. We test this by disabling Cypher query generation and supplying incorrect Cypher queries for each question, resulting in incorrect graph results.


\begin{table}[!ht]
\centering
\begin{tabular}{@{}lrr@{}}
\toprule
 \textbf{(in \%)} & \textbf{gpt-4o} & \textbf{gpt-4-turbo} \\
\midrule
Answer Denied & 65/69 (94.2) & 63/69 (91.3)  \\
Uncertain Answer & 1/69 (1.5) & 1/69 (1.5) \\
Full Answer & 3/69 (4.3) & 5/69 (7.3) \\
\bottomrule
\end{tabular}
\caption{Handling irrelevant information in graph responses.}
\label{tab:irrelevant-info}
\end{table}

The results in Table \ref{tab:irrelevant-info} show that both GPT-4-turbo and GPT-4-o can detect irrelevant information and correctly respond with "I don't know" in over 90\% of cases, demonstrating that the LLMs can reason and understand when the knowledge passed to them is not relevant. This highlights FactFinder's ability to enhance reliability by leveraging both structured and world knowledge.

Manual analysis revealed that in one case, the LLM expressed uncertainty with the phrase "The provided information does not mention ..." indicating potential inaccuracies. In a few cases, the models provided full answers despite unrelated KG results, especially for count, boolean, and lengthy responses, highlighting detection challenges in these scenarios.

\section{Conclusion}
This work demonstrates the value of integrating structured, factual knowledge into a user-friendly chat system, providing researchers with a reliable tool for answering scientific questions while minimizing hallucinations. We show that LLMs can generate valid Cypher queries to retrieve relevant data from a KG, informing accurate answers. The creation of such a robust system is crucial for enhancing research capabilities. Future work will focus on expanding the evaluation dataset, quantifying system uncertainty, and enabling access to multiple KGs, possibly through agent-based retrieval.

\bibliography{main-bib}

\appendix

\section{Appendix}
\label{sec:appendix}

\begin{figure*}[!htb]
\begin{center}
\includegraphics[width=0.92\textwidth]{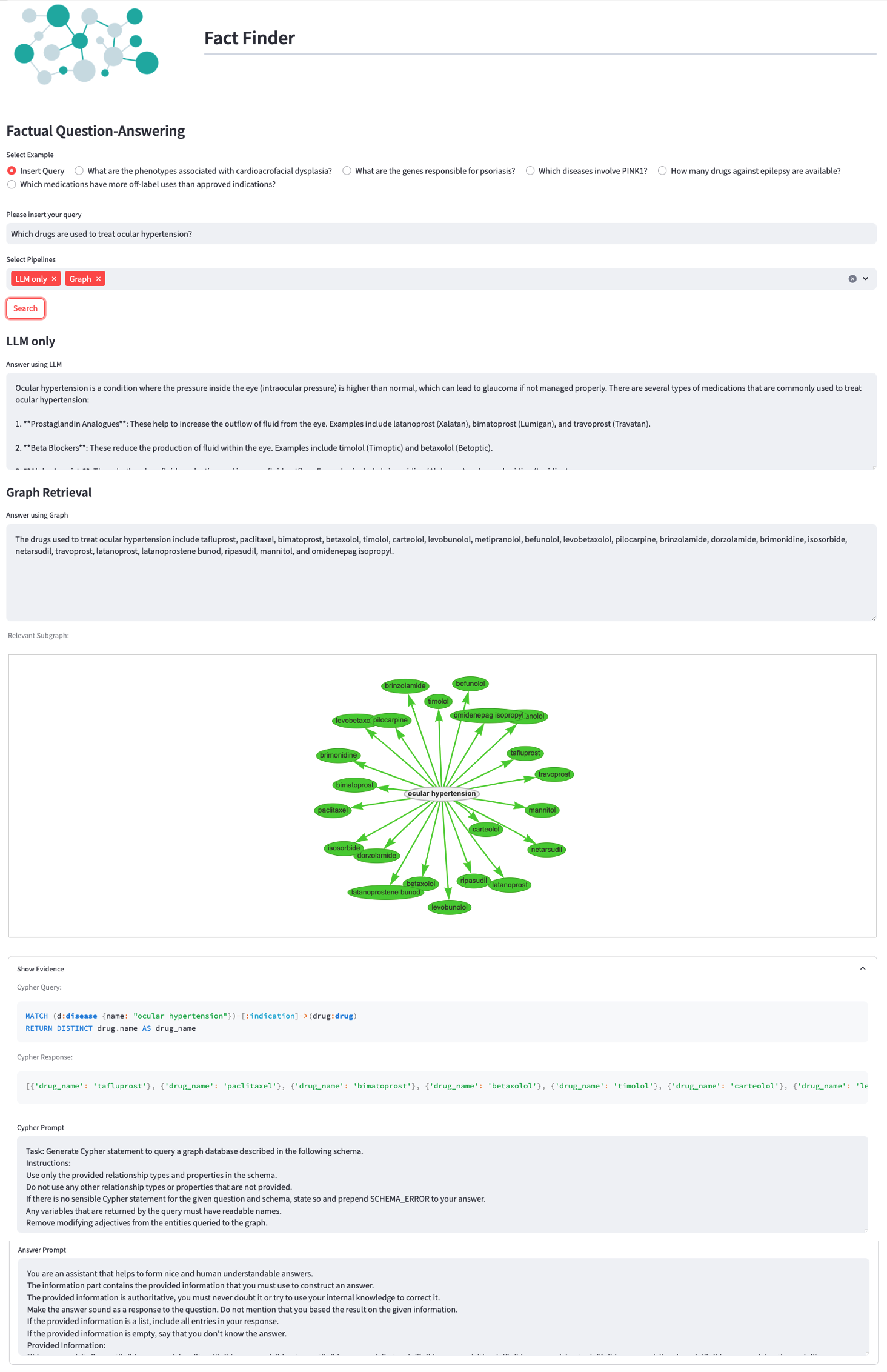}
\caption{User interface of Fact Finder for the question \textit{Which drugs are used to treat ocular hypertension?}. The answers of the standalone LLM and our graph-based hybrid system are compared as output. In addition, the relevant subgraph is displayed as evidence together with the generated Cypher query and the answer from the graph.}
\label{fig:ui}
\end{center}
\end{figure*}

\begin{figure*}[!htb]
\begin{center}
\includegraphics[width=0.98\textwidth]{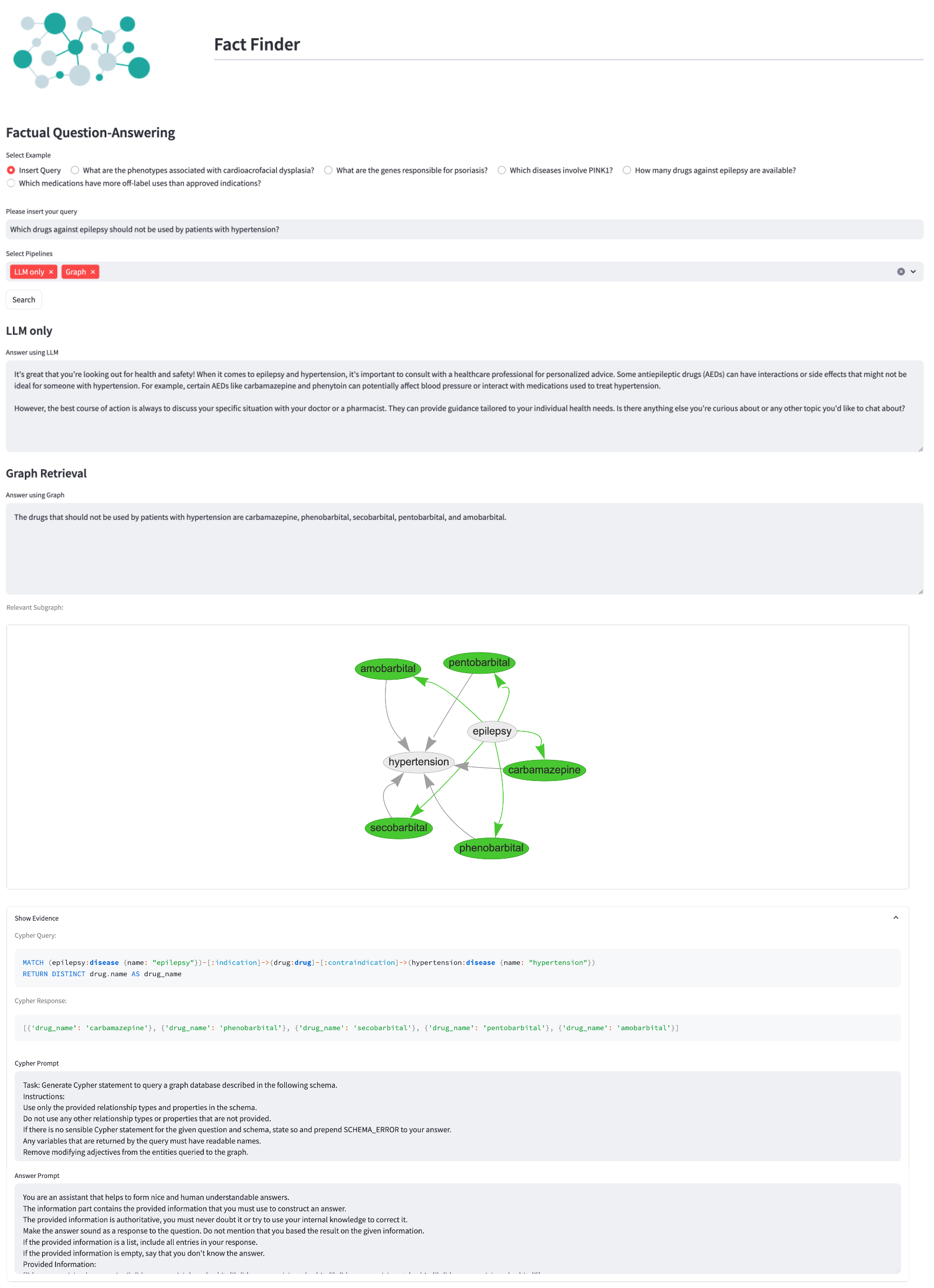}
\caption{User interface of Fact Finder for the question \textit{Which drugs against epilepsy should not be used by patients with hypertension?}.}
\label{fig:ui}
\end{center}
\end{figure*}

\end{document}